\documentclass[10pt,twocolumn,letterpaper]{article}
\usepackage{iccv}
\usepackage{times}
\usepackage{epsfig}
\usepackage{graphicx}
\usepackage{amsmath}
\usepackage{amssymb}

\usepackage{enumitem}
\usepackage{booktabs}
\usepackage{caption}
\usepackage{ifthen}
\usepackage[normalem]{ulem}
\usepackage[table, dvipsnames]{xcolor}
\usepackage{algpseudocode}
\usepackage{algorithm}

\newcommand{\amit}[2][]{%
    \ifthenelse{ \equal{#1}{} }
        {\textcolor{Dandelion}{(AA) #2}}
        {\textcolor{Dandelion}{(AA) \sout{#1} #2}}
}
\newcommand{\tamar}[2][]{%
    \ifthenelse{ \equal{#1}{} }
        {\textcolor{Orchid}{(TK) #2}}
        {\textcolor{Orchid}{(TK) \sout{#1} #2}}
}
\newcommand{\shai}[2][]{%
    \ifthenelse{ \equal{#1}{} }
        {\textcolor{WildStrawberry}{(SB) #2}}
        {\textcolor{WildStrawberry}{(SB) \sout{#1} #2}}
}
\usepackage[pagebackref=true,breaklinks=true,letterpaper=true,colorlinks,bookmarks=false]{hyperref}

\usepackage[capitalize]{cleveref}
\crefname{section}{Sec.}{Secs.}
\Crefname{section}{Section}{Sections}
\Crefname{table}{Table}{Tables}
\crefname{table}{Tab.}{Tabs.}

\iccvfinalcopy 

\def\iccvPaperID{6} 

\ificcvfinal\fi

\graphicspath{{./images/}}

\begin{document}

\title{\mbox{DeepCut: Unsupervised Segmentation using Graph Neural Networks Clustering}}

\author[ ]{Amit Aflalo}
\author[ ]{Shai Bagon}
\author[ ]{Tamar Kashti}
\author[ ]{Yonina Eldar}

\affil[ ]{\small Faculty of Mathematics and Computer Science, Weizmann Institute of Science}
\twocolumn[{
\renewcommand\twocolumn[1][]{#1}
\maketitle
\centering
\vspace*{-8mm   }
\includegraphics[width=17.5cm]{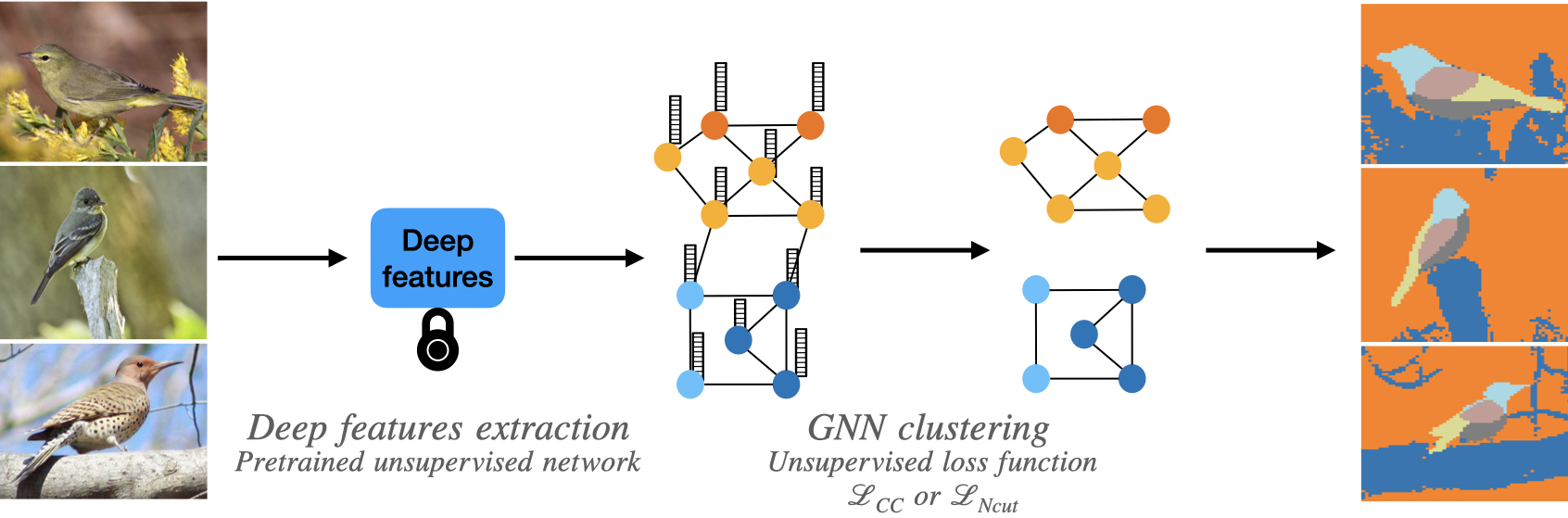}
\captionof{figure}{
\textbf{DeepCut:} We use graph neural networks with unsupervised losses from classical graph theory to solve various image segmentation tasks. Specifically, we employ deep features obtained from a pre-trained vision transformer to construct a graph representation for each image, and subsequently partition this graph to generate a segmentation. 
}
\label{fig:teaser}
 \vspace*{0.26cm}
}]

\maketitle

\begin{abstract}
Image segmentation is a fundamental task in computer vision.
Data annotation for training supervised methods can be labor-intensive, motivating unsupervised methods.
Current approaches often rely on extracting deep features from pre-trained networks to construct a graph, and classical clustering methods like $k$-means and normalized-cuts are then applied as a post-processing step. However, this approach reduces the high-dimensional information encoded in the features to pair-wise scalar affinities.
To address this limitation, this study introduces a lightweight Graph Neural Network (GNN) to replace classical clustering methods while optimizing for the same clustering objective function. Unlike existing methods, our GNN takes both the pair-wise affinities between local image features and the raw features as input. This direct connection between the raw features and the clustering objective enables us to implicitly perform classification of the clusters between different graphs, resulting in part semantic segmentation without the need for additional post-processing steps.
We demonstrate how classical clustering objectives can be formulated as self-supervised loss functions for training an image segmentation GNN. Furthermore, we employ the Correlation-Clustering (CC) objective to perform clustering without defining the number of clusters, allowing for $k$-less clustering.
We apply the proposed method for object localization, segmentation, and semantic part segmentation tasks, surpassing state-of-the-art performance on multiple benchmarks\footnote{Project page: \href{https://sampl-weizmann.github.io/DeepCut/}{https://sampl-weizmann.github.io/DeepCut/}}.
\end{abstract}

\section{Introduction}
\label{sec:intro}
Object localization and segmentation play crucial roles in various real-world applications, such as autonomous cars, robotics, and medical diagnosis. These tasks have been longstanding challenges in computer vision, and significant efforts have been invested in improving their accuracy.

Presently, state-of-the-art performance in these tasks is achieved using supervised Deep Neural Networks (DNNs). However, the limited availability of annotated data restricts the applicability of these methods. Data annotation is labor-intensive and costly, particularly in specialized fields like medical imaging, where domain experts are required for accurate annotations. Several solutions have been proposed to address this challenge, including leveraging color data, adding priors such as boundaries or scribbles \cite{lin2016scribblesup}, semi-supervised learning\cite{assran2021semi, kirillov2023segany}, weakly-supervised learning\cite{ren2020instance}, and more. However, these approaches have limitations since they still rely on some form of annotations or prior knowledge about the image structure.

An alternative approach to tackle this problem is to explore \emph{unsupervised} methods. Recent research on unsupervised Deep Neural Networks (DNNs) has yielded promising outcomes. For instance, self-DIstillation with No labels (DINO~\cite{caron2021emerging}) was employed to train Vision Transformers (ViTs), and the generated attention maps corresponded to semantic segments in the input image. The deep features extracted from these trained transformers demonstrated significant semantic meaning, facilitating their utilization in various visual tasks, such as object localization, segmentation, and semantic segmentation.\cite{amir2021deep, melas2022deep, wang2022self}.

Some recent work in this area has demonstrated promising results, especially using unsupervised techniques that combine deep features with classical graph theory for object localization and segmentation tasks \cite{melas2022deep, wang2022self}.
These methods are lightweight and rely on pre-trained unsupervised networks, unlike end-to-end approaches that require significant time and resources for training from scratch.

Our proposed method called \emph{DeepCut}, introduces an innovative approach using Graph Neural Networks (GNNs) combined with classical graph clustering objectives as a loss function. GNNs are specialized neural networks designed to process graph-structured data, and they have achieved remarkable results in various domains, including protein folding with AlphaFold \cite{jumper2021highly}, drug discovery \cite{you2018graph} and traffic prediction\cite{jiang2022graph}. This work employs GNNs for computer vision tasks such as object localization, segmentation, and semantic part segmentation.

A key advantage of our method is the direct utilization of deep features within the clustering process, in contrast to previous approaches \cite{melas2022deep, wang2022self} that discarded this valuable information and relied solely on correlations between features. This approach leads to improved object segmentation performance and enables semantic segmentation across multiple graphs, each corresponding to a separate image.
To further enhance segmentation, we adopt a two-stage approach that overcomes shortcomings of the objective functions overlooked by previous methods. We first separate foreground and background and then perform segmentation individually for each part, leading to more accurate results.
Moreover, we introduce a method to perform "k-less clustering," allowing data clustering without the need to predefine the number of clusters $k$, enhancing flexibility and adaptability.

\noindent{}Our contributions are:
\begin{itemize}[leftmargin=*]
\item Utilizing a lightweight GNN with classical clustering objectives as unsupervised loss functions for image segmentation, surpassing state-of-the-art performance at various unsupervised segmentation tasks (speed and accuracy).

\item[$\bullet$] Optimizing the correlation clustering objective for deep features clustering, which is not feasible with classical methods, thus achieving $k$-less clustering.

\item[$\bullet$] Performing semantic part segmentation on multiple images using test-time optimization. By applying the method to each image separately, we eliminate the need for any post-processing steps required by previous methods.
\end{itemize}


\section{Background}
In the era prior to the deep-learning surge, numerous classical approaches adopted quantitative criteria for segmentation based on principles from graph theory. These methods involved representing affinities between image regions as a graph and associating various image partitions with cuts in that graph (e.g., \cite{shi2000normalized, bansal2004correlation, bagon2011large}). The quality of image segments was determined by the "optimality" of these cuts. Remarkably, these techniques operated without any supervision, relying solely on the provided affinities between image regions.

Different methods have defined optimal cuts in various ways, each presenting advantages. We provide a brief overview of the relevant approaches.

 \subsection{Graph Clustering}\label{c_func}
 We leverage two graph clustering functionals from classical graph theory in our work, \emph{normalized  cut}\cite{shi2000normalized} and \emph{correlation  clustering}\cite{bansal2004correlation}. 
 
\paragraph{Notations}
Let $G = (V,E)$ be an undirected graph induced by an image. Each node represents an image region, and the weights $w_{ij}$ represent the affinity between image regions $i$ and $j$, $i,j=1\ldots n$. Let $W$ be an $n\times n$ matrix whose entries are $w_{ij}$. 

Our goal is to partition this graph into $k$ disjoint sets $A_1, A_2 ... A_k$
such that $\cup_i A_i = \mathcal{V}$ and  $\forall_{j\ne i} A_i \cap A_j = \emptyset$.
This partition can be expressed as a binary matrix $S\in\left\{0,1\right\}^{n\times k}$ where $S_{ic}=1$ iff $i\in A_c$.

\paragraph{Normalized Cut (N-cut)~\cite{shi2000normalized}}
A good partition is defined as one that maximizes the number of within-group
connections, and minimizes the number of between-group connections. The number of between-group connections 
can be computed as the total weight of edges removed and described in graph theory as a \emph{cut}:

\begin{equation}
  cut(A,B) = \sum_{u \in A, v \in B}w(u , v).
  \label{eq:min_cut}
\end{equation}

Where $A, B$ are parts of two-way partition of $G$. This objective is formulated by the \emph{normalized cut} (N-cut) functional:
\begin{equation}
  Ncut(A,B) = \frac{cut(A, B)}{assoc(A,\mathcal{V})} + \frac{cut(A, B)}{assoc(B,\mathcal{V})},
  \label{eq:n_cut}
\end{equation}
where $assoc(A,\mathcal{V})\!=\!\sum_{i \in A, j \in \mathcal{V}} w_{ij}$ is the total affinities connecting nodes of $A$ to all nodes in the graph.
The N-cut formulation can be easily extended to $K>2$ segments. 
Shi and Malik \cite{shi2000normalized} also suggested an approximated solution to \cref{eq:n_cut} using spectral methods, known as \emph{spectral clustering}, where the spectrum (eigenvalues) of a graph Laplacian matrix is used to get approximated solution to the N-cut problem.

\paragraph{Correlation Clustering (CC)~\cite{bansal2004correlation}}
When the graph, derived from an image, contains \emph{both negative and positive} affinities, N-cut is no longer applicable. 
In this case, a good image segment may be one that maximizes the \emph{positive} affinities inside the segment and the \emph{negative} ones across segments~\cite{bansal2004correlation}. 
This objective can be formulated by the \emph{correlation clustering} (CC) functional. 
Given a matrix $W$, an optimal partition $U$ minimizes:
\begin{equation}
CC(S) = - \sum_{ij}W_{ij} \sum_{c} S_{ic}S_{jc}.
 \label{eq:jj}
\end{equation}
Correlation clustering utilizes intra-cluster disagreement (repulsion) to \emph{automatically} deduce the number of clusters $k$~\cite{bansal2004correlation}  so that it does not need to know $k$ in advance. Further theoretical analysis on this property can be found in~\cite{bagon2011large}, along with classical optimization algorithms applying CC to image segmentation.

\subsection{Graph Neural Networks}
GNN (Graph Neural Network) is a category of neural networks designed to process graph-structured data directly. A GNN layer comprises two fundamental operations: \textbf{message passing} and \textbf{aggregation}.
\textbf{Message passing} involves gathering information from the neighbors of each node and is applied to all nodes in the graph. The exchanged information can include a combination of node features, edge features, or any other data embedded in the graph.
\textbf{Aggregation} refers to fusing all the messages obtained during the message passing phase into one message that updates the current node's state.

In our approach, we utilize a specific type of GNN called the Graph Convolutional Network (GCN) \cite{kipf2016semi}. In a GCN layer $\ell$, each node $h_v$ is processed as follows:
\begin{equation}
{h}_v^{(\ell+1)} = \sum_{u\in \mathcal{N}(v)} \mathbf{\Theta} \frac{{h_u}^{(\ell)}}{|\mathcal{N}(v)|},
  \label{eq:gcn}
 \end{equation}
 where $\mathbf{\Theta}$ is a matrix of trainable parameters. $\mathcal{N}(v)$ denotes the neighbours of a node $h_v$, and $|\mathcal{N}(v)|$ is the number of neighbours. The GCN aggregation is performed through a summation operator, with the term inside the summation representing the message-passing operation. The objective during training is to optimize the network parameters $\mathbf{\Theta}$ in a way that generates meaningful messages to be passed among nodes, optimizing the loss function.

\section{Method}\label{sec:method}
In our approach, we process each image through the pre-trained network to extract deep features, which are then used for clustering. Subsequently, we construct a weighted graph based on these features. Employing a lightweight Graph Neural Network (GNN), we optimize our unsupervised graph partitioning loss functions (either $\mathcal{L}{NCut}$ or $\mathcal{L}{CC}$) separately for each image graph.

We adopt this methodology to achieve unsupervised object localization, segmentation, and semantic part segmentation.
  
\begin{figure*}[ht]
\includegraphics[width=17.5cm]{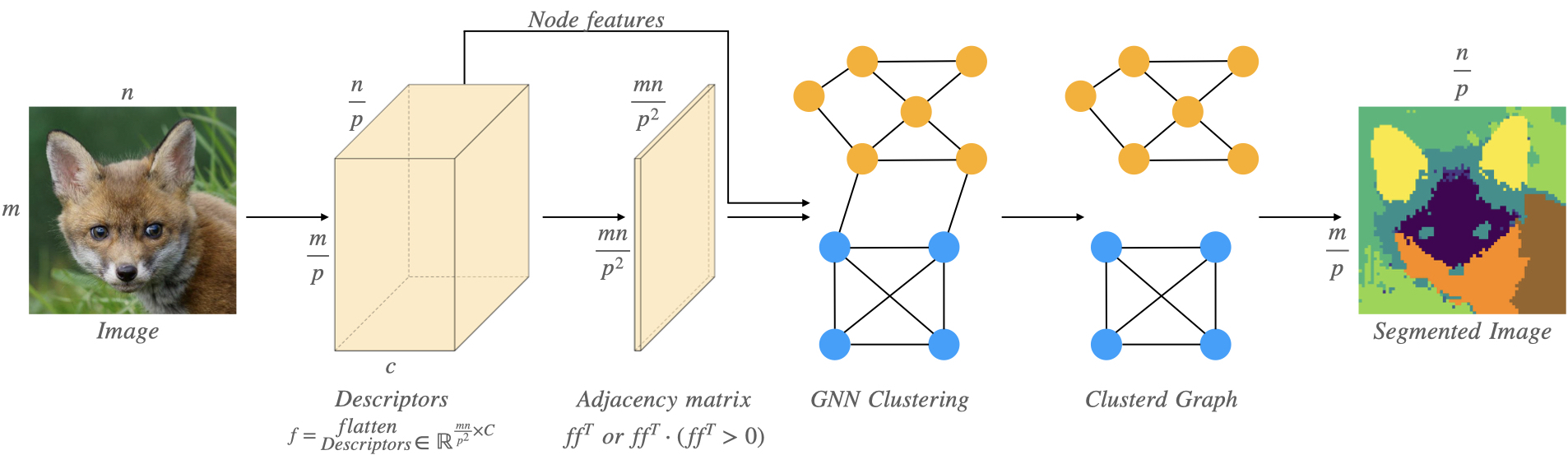}
\caption{
\textbf{Method overview:} After extracting deep features from a pretrained ViT model, we construct a similarity matrix based on the patch-wise feature similarities, which becomes our adjacency matrix. We build a graph using this adjacency matrix and the deep features as node features. Next, we train a lightweight GNN using unsupervised graph partitioning loss functions (\cref{c_func}) to partition the graph into $k$ distinct clusters, which can be used for various downstream tasks.
}
\label{fig:arch}
\end{figure*}

\subsection{From Deep Features to Graphs}\label{sec:deep_to}
As shown in \Cref{fig:arch}, given an image $M$ of size $m \times n$ and $d$ channels, we pass it through a transformer T. 
The transformer divides the image into $\frac{mn}{p^2}$ patches,
where $p$ is the patch size of transformer T. To extract the transformer's internal representation for each patch, we utilize the \emph{key} token from the last layer, as it has demonstrated superior performance across various tasks \cite{amir2021deep, melas2022deep}. The output is a feature vector $f_{(\frac{mn}{p^2} \times c)}$, where $c$ represents the token embedding dimension, containing all the extracted features from the different patches.

Consider a weighted graph $G = (V, E)$ with $W$ as the weight matrix. We construct a patch-wise correlation matrix from the features obtained by the Vision Transformer:
\begin{equation}
	W = ff^T \in \mathbb{R}^{\frac{mn}{p^2} \times \frac{mn}{p^2}}.
  \label{eq:cc_adj}
\end{equation}
In correlation clustering, the negative weights (representing reputations) carry valuable information utilized by the clustering objective. However, the normalized cut objective only accepts positive weights. As a result, we threshold at zero:

\begin{equation}
	W = ff^T \cdot(ff^T > 0) \in \mathbb{R}^{\frac{mn}{p^2} \times \frac{mn}{p^2}}.
  \label{eq:norm_adj}
\end{equation}

We introduce a hyper-parameter called \emph{k sensitivity} denoted by $\alpha$ to adapt the cluster choosing process in correlation clustering. Since the number of clusters cannot be directly chosen in correlation clustering, this parameter allows us to control the sensitivity of the process, where a higher value of $\alpha$ corresponds to more clusters. We enforce this adjustment in the following manner:

\begin{equation}
	W = ff^T - \frac{max(ff^T)}{\alpha}.
  \label{eq:cor_sen}
\end{equation}
where $\alpha \in [1, \inf)$. Lower $\alpha$ value corresponds to higher repulsion forces between nodes (negative weights in $W$) and thus higher cluster count, as correlation clustering maximizes negative affinities between segments in the graph.


\subsection{Graph Neural Network Clustering}\label{sec:gnn_clus}
Let \textbf{$\hat{N}$} be a node feature matrix obtained by applying one or more layers of GNN convolution on a graph $G$ with an adjacency matrix $W$. In our case, we use a one-layer GCN and construct the graph using the patch-wise correlation matrix from ViT obtained features (\cref{eq:cc_adj}, \cref{eq:norm_adj}, \cref{eq:cor_sen}).
Let \textbf{S} be the output of a Multi-Layer Perception (MLP) with a softmax function applied on \textbf{$\hat{N}$}:
\begin{equation}
\begin{gathered}
	\hat{N} = GNN(N , W; \Theta_{GNN}), \\ S = MLP({\hat{N}; \Theta_{MLP}}),
\end{gathered}
  \label{eq:gnn_cut}
\end{equation}
where $\Theta_{MLP}$ and $\Theta_{GNN}$ are trainable parameters. The GNN output $S$, is the cluster assignment matrix containing vectors representing the node's probability of belonging to a particular cluster.

The GNN is optimized using either the normalized-cut relaxation proposed in \cite{bianchi2020spectral} or our newly proposed method with the correlation clustering objective as the loss function.

The loss function in the case of normalized cut is:
\begin{equation}
	\mathcal{L}_{NCuts} = \frac{Tr(S^TWS)}{Tr(S^TDS)} + \left\lVert \frac{S^TS}{\left\lVert  S^TS \right\rVert _F}  - \frac{\mathbb{I}_K}{\sqrt{K}}  \right\rVert_F,
  \label{eq:cut}
\end{equation}
where $D=diag(\sum_{j}W_{i,j})$ is the row-wise sum diagonal matrix of $W$. $K$ denotes the number of disjoint sets we aim to partition the graph into, and $\mathbb{I}_K$ is the identity matrix.
The first term of the objective function promotes the clustering of strongly connected components together, while the second term encourages the cluster assignments to be orthogonal and have similar sizes.

The loss function for correlation clustering \cite{bansal2004correlation} is:
\begin{equation}
	\mathcal{L}_{CC} = -Tr(WSS^T).
  \label{eq:cc}
\end{equation}

This term promotes intra-cluster agreement while encouraging repulsion (negative affinities) between clusters.

$W$ is defined as \cref{eq:norm_adj} and \cref{eq:cor_sen} for the N-cut and CC loss, respectively. To control the sensitivity of the clustering process for CC, we utilize a k-sensitivity value, as explained in \cref{sec:k_les_l}.
 
\subsection{Graph Neural Network Segmentation}

In \cref{sec:deep_to}, we propose constructing a graph from deep features extracted from unsupervised trained ViTs, where each node represents a patch of the original image. The nodes in the graph are then clustered into disjoint sets, representing different image segments. For this clustering process, we employ graph neural network clustering as described in \cref{sec:gnn_clus}, utilizing either the CC or N-cut losses. However, unlike previous approaches that used the N-cut loss defined strictly for positive weights \cite{bianchi2020spectral, melas2022deep}, we advocate using the correlation clustering functional as a loss. Correlation clustering enables us to utilize negative weights for graph building (see \cref{eq:cc_adj} and \cref{eq:cor_sen}), facilitating clustering without predefining the number of clusters.

Additionally, we incorporate deep features as node features for graph building, which is absent in previous methods that solely used the correlation matrix of the deep features \cite{bianchi2020spectral, melas2022deep, wang2022self}. As a result, our method allows for feature classification while clustering and implicitly facilitates semantic part segmentation without necessitating post-processing steps, as seen in previous methods \cite{melas2022deep}.

Our approach is versatile and applicable to various image partitioning-related tasks, including:

 
\paragraph{Object Localization}
Object localization involves identifying the primary object in an image and enclosing it with a bounding box. To perform localization, we follow the steps below:
(1)~Use our GCN clustering method with $k = 2$. (2)~Examine the edges of the clustered image and identify the cluster that appears on more than two edges as the background, while the other cluster becomes our main object. (3)~Apply a bounding box around the identified main object.

  \begin{figure}[t]
\includegraphics[width= \linewidth]{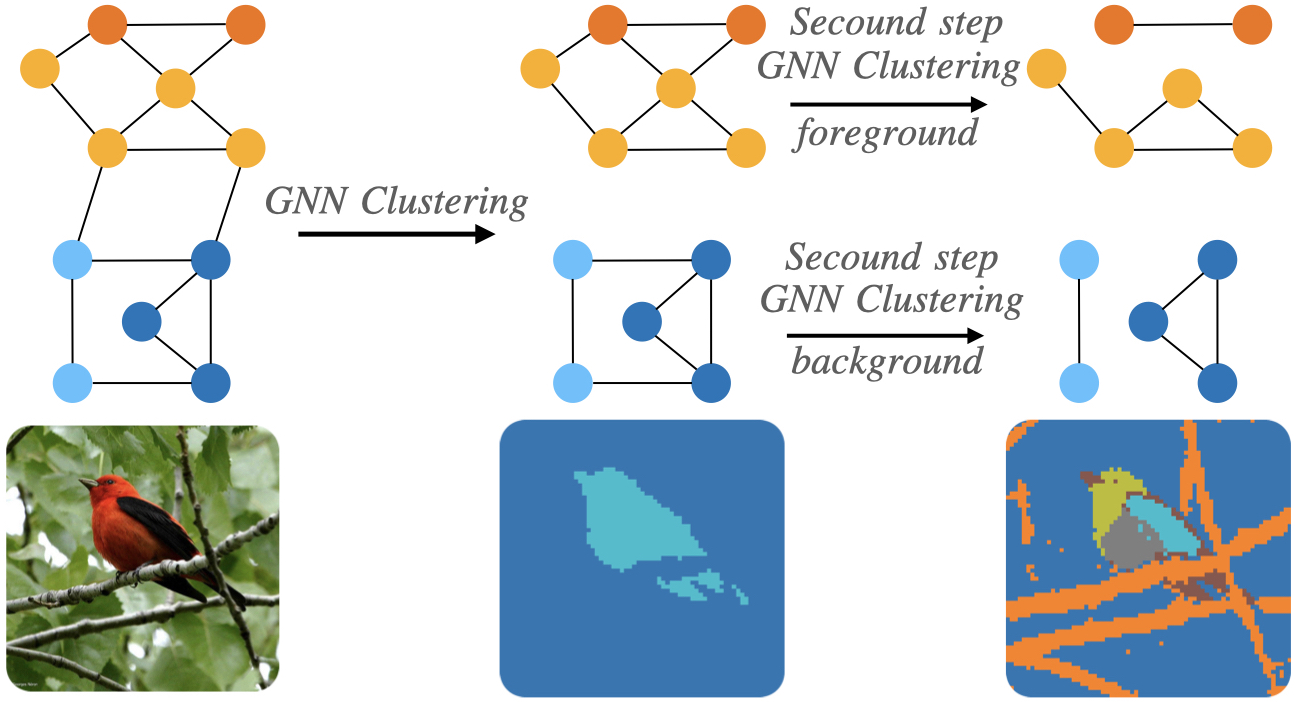}
\caption{
\textbf{Proposed two-stage clustering:} First, we cluster the image into two disjoined sets,  then apply clustering to the foreground and background separately.
}
\label{fig:hier}
 \end{figure}


\paragraph{Object Segmentation}
Object segmentation involves the separation of the foreground object in an image, commonly known as foreground-background segmentation. The method employed for object segmentation is identical to object localization, with the inclusion of stage (3), where the bounding box is applied.
 
\paragraph{Semantic Part Segmentation} 

DeepCut achieves semantic part segmentation through a test-time optimization paradigm, where the model is sequentially exposed to each image, optimizing the model weights based on the previous image. This means the model does not require training on all images beforehand, eliminating the need for co-segmentation or post-processing steps.

This advantage is derived from our approach to learning deep features with GNN, which differs from previous methods \cite{bianchi2020spectral, wang2022self} that solely rely on correlations and discard the high-dimensional data of deep features. By leveraging the intricate semantic information embedded within deep features, our model implicitly performs the classification of the clusters between different image graphs, enabling semantic part segmentation without the need for explicit post-processing or additional training steps.

Our segmentation process involves two steps, as shown in Figure \ref{fig:hier}: foreground-background segmentation (k=2) followed by semantic part segmentation on the foreground object (k=4). This two-stage process addresses the bias of the clustering functions towards larger clusters (e.g., background-foreground), which limits the level of detail in foreground object segmentation. The exact process can also be applied to improve background segmentation, as depicted in Figure \ref{fig:teaser} and Figure \ref{fig:main}.


\begin{table}
  \centering
  \begin{tabular}{@{}lc c c@{}}
    \toprule
    Method & VOC-07 & VOC-12 &COCO-20k\\
    \midrule
    Selective Search\cite{uijlings2013selective} & 18.8 & 20.9 & 6.0 \\
    EdgeBoxes\cite{zitnick2014edge} & 31.1 & 31.6 & 28.8 \\
    DINO-[CLS]\cite{caron2021emerging} & 45.8 & 46.2 & 42.1 \\
    LOST\cite{simeoni2021localizing} & 61.9 & 64.0 &  50.7\\
    Spectral Methods\cite{melas2022deep} & 62.7 & 66.4 &  52.2\\
    TokenCut\cite{wang2022self} & 68.8 & 72.1 &  58.8\\
    \midrule
    DeepCut:  CC loss&68.8 & 67.9 & 57.6\\
    DeepCut:  N-cut loss&\textbf{69.8} & \textbf{72.2} & \textbf{61.6}\\
    \bottomrule
  \end{tabular}
  \caption{\textbf{Object localization results.} CorLoc metric (percentage of images with $IOU > 0.5$). 
  CC and N-cut denotes correlation clustering and Normalized Cut respectively.}
  \label{tab:obdet}
\end{table}
 
\section{Training and performance}
For all experiments, We use DINO \cite{caron2021emerging} trained ViT-S/8 transformer for feature extraction, with pre-trained weights from the DINO paper authors (trained on ImageNet\cite{ILSVRC15}). 
 \textbf{No training is conducted on the tested datasets.}
We employ a test-time training paradigm for all experiments, where each image is trained using a proposed graph neural network (GNN) segmentation approach for ten epochs. Since the proposed losses are unsupervised and involve solving an optimization problem, generalizing the model to the entire dataset does not improve accuracy. This training method achieves a performance (speed) that is \textbf{more than two times} that of the current state-of-the-art TokenCut\cite{wang2022self} on the same hardware. DeepCut efficiency stems from its lightweight architecture, consisting of only 30k trainable parameters. Implementation details and performance analyses are provided in the supplementary material.
 
\section{Results}\label{sec:results}
Our method is evaluated on three unsupervised tasks: single object localization, single object segmentation, and semantic part segmentation. We compare our approach with other unsupervised methods published on these tasks using widely used benchmarks. The results are presented for both the normalized-cut and correlation clustering GNN objectives.


\begin{table}
  \centering
  \begin{tabular}{@{}lc c c@{}}
    \toprule
    Method & CUB &  DUTS & ECSSD\\
    \midrule
    OneGAN\cite{benny2020onegan} & 55.5 & -  & -\\
    Voynov et al.\cite{voynov2021object} & 68.3 & 49.8& - \\
    Spectral Methods\cite{melas2022deep} & 76.9 &  51.4 & 73.3\\
    TokenCut\cite{wang2022self} & - &  57.6& 71.2\\
    \midrule
    DeepCut: CC loss&77.7 & 56.0 & 73.4\\
    DeepCut: N-cut loss&\textbf{78.2} & \textbf{59.5} & \textbf{74.6}\\
    \bottomrule
  \end{tabular}
  \caption{\textbf{Single object segmentation results.} mIOU (mean intersection-over-union) CC denotes correlation clustering and N-cut Normalized Cut.}
  \label{tab:obseg}
\end{table}


\subsection{Object Localization}
In  \Cref{tab:obdet}, We evaluate our unsupervised object localization performance on three datasets: PASCAL VOC 2007 \cite{pascal-voc-2007}, PASCAL VOC 2012 \cite{pascal-voc-2012}, 
and COCO20K (20k images chosen from MS-COCO dataset\cite{lin2014microsoft} introduced in previous work\cite{vo2020toward}).
We compare our unsupervised approach to state-of-the-art unsupervised methods.
We report our results in the Correct Localization metric (CorLoc), defined as the percentage of images whose intersection-over-union with the ground truth label is greater than 50\%. 
DeepCut with N-cut losses gives the best results on all data sets, and DeepCut with correlation clustering loss surpasses all previous methods except TokenCut\cite{wang2022self}.

\subsection{Single Object Segmentation}
\begin{figure}[t]
\includegraphics[width= \linewidth]{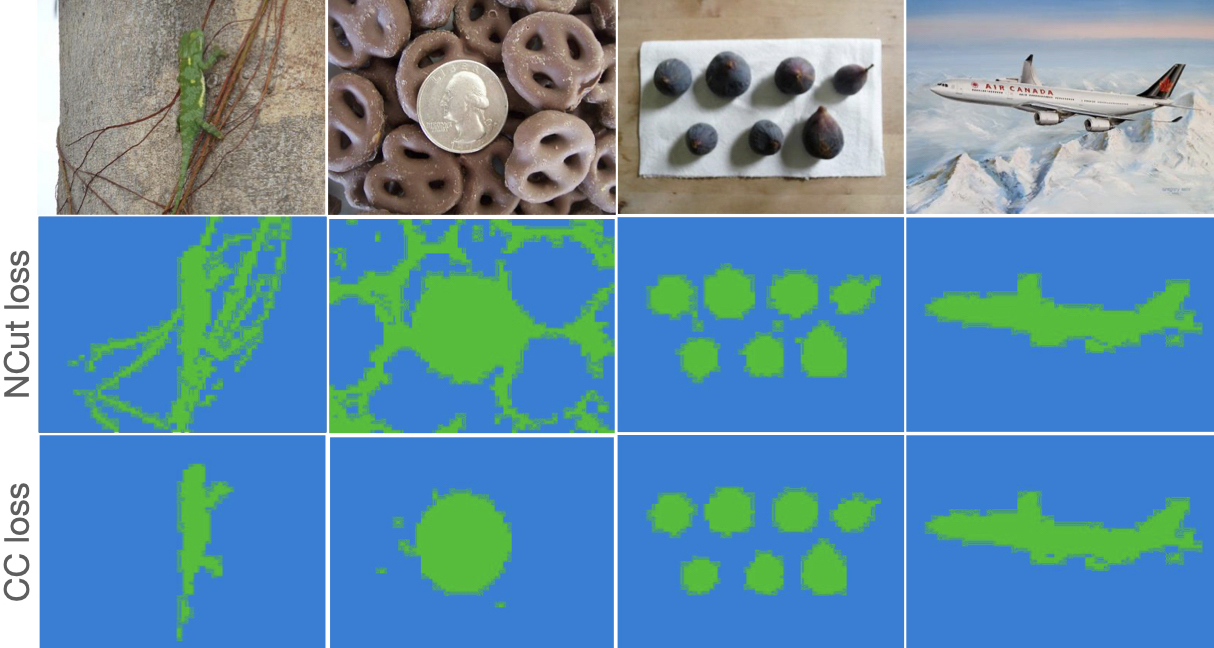}
\caption{
\textbf{DeepCut segmentation: N-cut vs. CC losses.}  
Top: original image, middle: DeepCut with N-cut loss, bottom: DeepCut with correlation clustering loss. Note that for images with complex background DeepCut with CC loss outperforms DeepCut with N-cut loss.}
\label{fig:seg_ex}
\end{figure}

In \Cref{tab:obseg}, We evaluate our unsupervised single object segmentation performance on three datasets: CUB (widely-used dataset of birds for fine-grained visual categorization task)\cite{cub_2011}, DUTS (the largest saliency detection benchmark)\cite{wang2017} and ECSSD (Extended Complex Scene Saliency Dataset)\cite{yan2013hierarchical}.
We report our results in mean intersection-over-union (mIoU). DeepCut with N-cut losses archives the best results on all data sets. Visual comparison of segmentation with our two losses is presented in \Cref{fig:seg_ex}; note how the CC loss segments the foreground better than the Ncut loss when background for specific images that include different objects and shadows. As this is not the case in most images, the N-cut usually outperforms the CC loss.

\begin{table}
  \centering
  \begin{tabular}{@{}lc c@{}}
    \toprule
    Method & NMI  & ARI\\
    \midrule
    SCOPS\cite{hung2019scops} (model) & 24.4 & 7.1\\
    Huang and Li\cite{huang2020interpretable} &  26.1& 13.2 \\
    Choudhury et al.\cite{choudhury2021unsupervised} & 43.5 &   19.6\\
    \midrule
    DFF\cite{collins2018deep} &  25.9 & 12.4\\
    Amir et al.\cite{collins2018deep} &  38.9 & 16.1\\
    \midrule
    DeepCut: N-cut loss&\textbf{43.9} &  \textbf{20.2}\\
    \bottomrule
  \end{tabular}
  \caption{\textbf{Semantic part segmentation results.} ARI and NMI over the entire CUB-200 dataset are used to evaluate cluster quality. Predictions were performed with $k=4$
  with our method using the N-cut objective. First three methods use ground truth foreground masks as supervision.}
  \label{tab:semseg}
\end{table}

\begin{figure}[t]
\includegraphics[width= \linewidth]{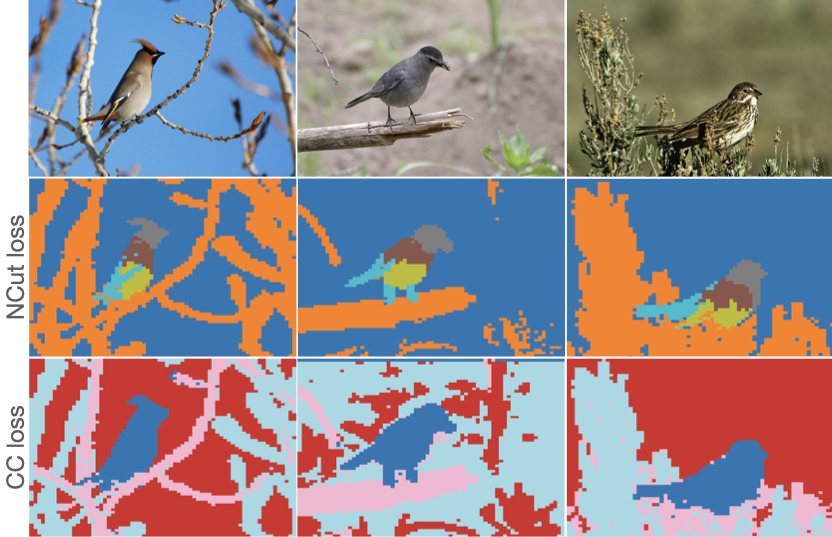}
\caption{
\textbf{Semantic part segmentation:  N-cut vs. CC losses.} Top: original image, middle: normalized-cut loss with our proposed two-step clustering; bottom: correlation clustering with one-step $k$-less clustering.}
\label{fig:main}
\end{figure}

\subsection{Semantic Part Segmentation}

\begin{figure}[t]
\includegraphics[width= \linewidth]{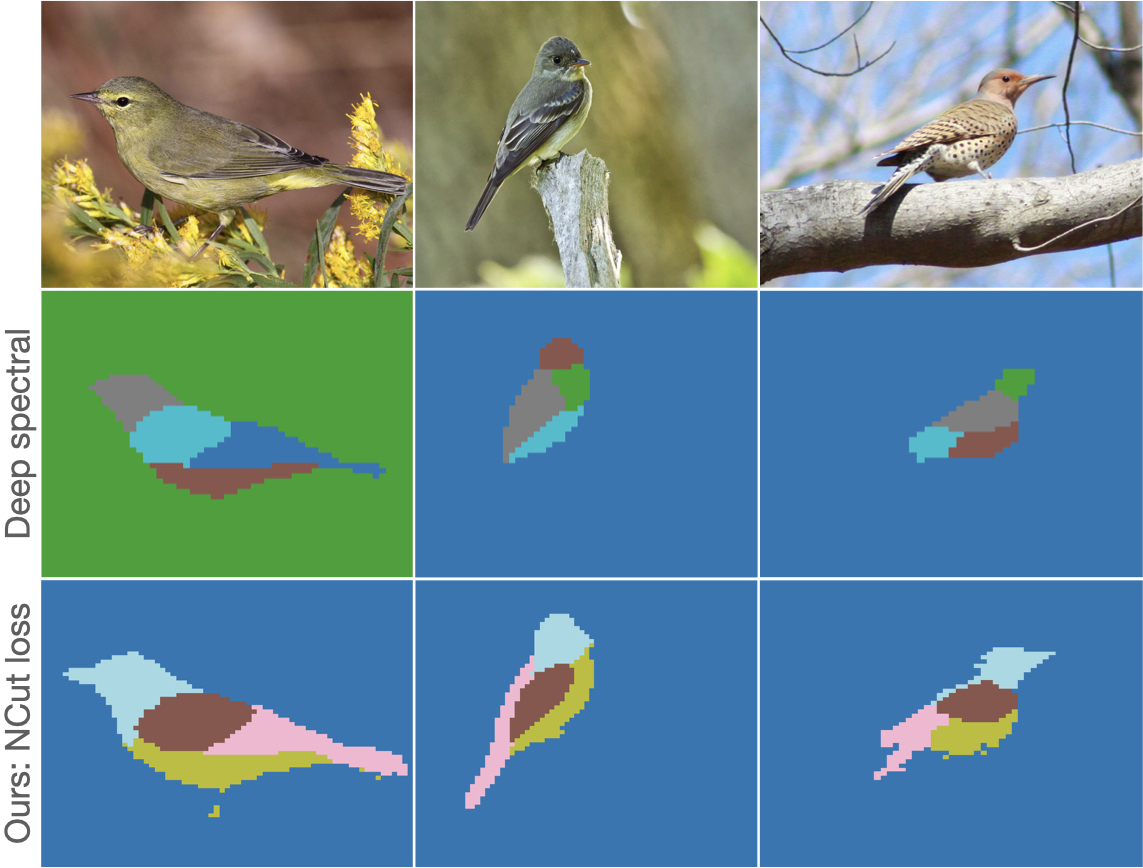}
\caption{
\textbf{Semantic part segmentation. deep spectral method vs. DeepCut with N-cut loss.} Top: original image; middle: deep spectral method\cite{melas2022deep}; bottom: our segmentation with N-cut loss. The deep spectral method failed to preform semantic segmentation across all three images.}
\label{fig:comp}
\end{figure}

In~\Cref{tab:semseg}, 
We evaluate our approach on the CUB dataset \cite{cub_2011} and report results using the Normalized Mutual Info score (NMI) and Adjusted Rand Index score (ARI) on the entire test set. A comparison between our technique and deep spectral method that uses classical graph theory for deep features-based segmentation\cite{bianchi2020spectral} is presented in \Cref{fig:comp}. As seen at~\Cref{tab:semseg}, DeepCut with normalized-cut loss surpasses all other methods, including the top three methods\cite{hung2019scops,huang2020interpretable,choudhury2021unsupervised} that uses ground-truth foreground masks as supervision.

\section{Deep Features Selection}\label{sec:deep_features}
Our work and previous unsupervised segmentation methods heavily rely on the correlation between deep features from different patches. However, in this study, we proposed a novel approach by utilizing the negative correlations (correlation clustering loss) that were discarded in previous works. By doing so, we leverage all available information to enhance performance in unsupervised segmentation. It is crucial to recognize that the performance of unsupervised segmentation methods that rely on deep features depends on the quality of these features.

Furthermore, no universal solution fits all scenarios, as different methods require specific types of information to achieve optimal results. For instance, our paper presents two techniques—one based on correlation clustering and the other on normalized cut. Correlation clustering depends on both positive and negative correlations between features, whereas normalized cut only accepts positive correlations.

In \Cref{tab:feats}, we compare three unsupervised ViT training approaches: DINO\cite{caron2021emerging}, MoCo\cite{chen2021empirical}, and MAE\cite{he2022masked}. We observe that correlation clustering performs best with a method that supplies deep features with the largest amount of negative correlation between features (DINO $33.6\%$). We also observe that even a small amount of negative correlation can be counterproductive, even with normalized cut (MAE $5.4\%$) that relies solely on positive correlation.

Considering these factors, it becomes essential to identify the best combinations of unsupervised training methods and segmentation techniques. This insight enables us to improve unsupervised training methods, allowing us to extract more valuable information for segmentation purposes.

\begin{table}
  \centering
  \begin{tabular}{@{}lc c c@{}}
    \toprule
    Pretraining  & DeepCut & DeepCut  &Negative   \\
     &  N-cut & CC & percentage  \\

        \midrule
    DINO\cite{caron2021emerging}  & 59.4 & 59.8 & 33.6\\
    MoCo-v3\cite{chen2021empirical}  & 62.3 & 43.1 & 20\\
    MAE \cite{he2022masked}  & 47.2 &  31.2 & 5.4\\

    \bottomrule
  \end{tabular}
  \caption{\textbf{Pretraining.} Object-localization performance on PASCAL VOC07\cite{pascal-voc-2007}
  with different unsupervised training methods for acquiring deep features. All methods use ViT-B-16 architecture, and were trained on ImageNet\cite{ILSVRC15}.
  We provide mIOU (mean Intersection-Over-Union) for correlation clustering (CC) and normalized-cut (N-cut) objective.
  We also provide the mean percentage (across all the dataset) of negative weights in the corresponding affinity matrix. Note the correlation: the higher the percentage of negative weights of the transformer, the better the mIOU of the DeepCut with CC loss; for DINO DeepCut CC loss outperforms N-cut loss for this dataset.
  }
  \label{tab:feats}
\end{table}

 \begin{table*}
  \centering
  \begin{tabular}{@{}l c c c c @{}}
    \toprule
    Number  of& 3 classes &4 classes&5 classes&\\
     \ \ \ classes  &                 &   \\

    \midrule
    Connected Components  & 33.3 $\pm 0$ & 25.0 $\pm 0$ & 20.0 $\pm 0$\\
    Spectral Clustering& 85.5 $\pm 14.1$ & 77.9 $\pm 6.6$ & 82.8 $\pm 6.9$ \\
    DeepCut: cc loss& 98.3 $\pm 0.9$  & 97.1 $\pm 1.7$ & 99.27 $\pm 0.6$  \\

    \bottomrule
  \end{tabular}
  \caption{\textbf{$k$-less clustering.} We apply DeepCut with correlation clustering loss on the Fashion Product Images Dataset\cite{Fashion_dataset}, using a subset of images from 5 classes: Top-wear, Shoes, Bags, Eye-wear, and Belts. We employ our $k$-less method and report the results as classification purity. The experiments are conducted with \emph{k sensitivity = 3} as defined in \cref{eq:cor_sen}. The presented results are the mean and standard deviation (\emph{std}) obtained from multiple experiments.
  }
  \label{tab:k_less}
\end{table*}


\section{$k$-less Clustering}\label{sec:k_les_l}
In this section, we explore clustering, using image classification as an example. Traditional clustering methods usually require a predefined number of clusters (or classes) $k$ for classifying data, which can be a disadvantage as it demands prior knowledge about the data. Our DeepCut method introduces a GNN model with correlation clustering loss, enabling $k$-less clustering, where the number of clusters is derived from the data. To demonstrate this valuable property, we use the Fashion Product Images Dataset\cite{Fashion_dataset}, sampling a subset of 500 images from 5 classes: Top-wear, Shoes, Bags, Eye-wear, and Belts.

We conducted three classification experiments, each repeated ten times: one with three classes, one with four, and one with all classes. We employ our GNN approach with correlation clustering loss in each experiment to cluster the images into different groups. We set the parameter \emph{$\alpha$} (\emph{k sensitivity} defined in \cref{eq:cor_sen}) to 3, using class tokens extracted from the unsupervised trained DINO\cite{caron2021emerging} ViT base transformer with a patch size of 16 as our features. We report the clustering result as classification purity in \cref{tab:k_less}. To validate the algorithm's robustness, we apply class permutation and random subset sampling of each class. 

\begin{figure}[t]
\includegraphics[width= \linewidth]{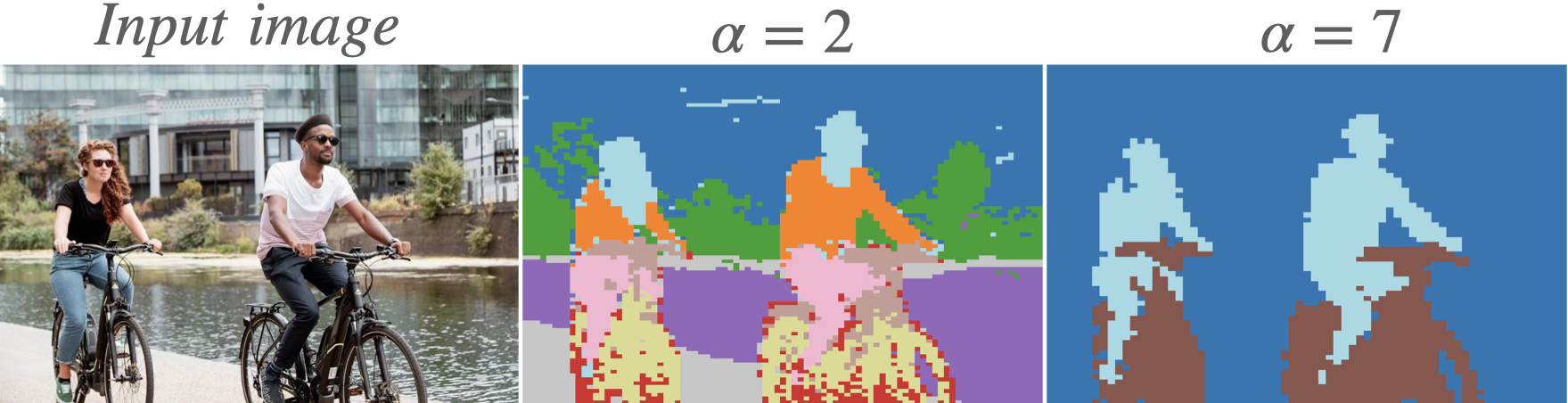}
\caption{
\textbf{k-sensitivity.} The CC segmentation results demonstrate that similar objects, such as people and bikes, cluster together consistently across different $\alpha$ values. Smaller $\alpha$ values lead to a finer partition of objects, but even then, CC still assigns similar objects to the same clusters.
}
\label{fig:rev}
 \end{figure}

The versatility of correlation clustering's $k$-less nature extends to segmentation tasks, as illustrated in \Cref{fig:main}. We compare our proposed DeepCut method with two baselines: connected component and spectral clustering. For spectral clustering, we use eigen-gap\cite{von2007tutorial} to select the number of clusters. The connected components method proves ineffective, clustering all images into the same group for all experiments, resulting in zero standard deviation (STD). On the other hand, spectral clustering performs better but exhibits high variance. In contrast, our DeepCut with correlation clustering loss achieves the highest accuracy and lowest variance. This powerful $k$-less property of correlation clustering is demonstrated through an image clustering example in terms of quantitative results. Furthermore, it can be effectively applied to image segmentation, as depicted in \cref{fig:rev}.

\paragraph{Choosing $\alpha$:}
The CC functional determines the number of clusters by considering repulsion forces between clusters, as explained in \cref{sec:method}. These forces are calculated based on the amount of negative affinities in the adjacency matrix defined at \cref{eq:cor_sen}. The affinities can vary when using differently trained ViT models (as observed in \cref{tab:feats}) or different datasets. To address this issue, we propose introducing a k-sensitivity variable $\alpha$ (as defined in \cref{eq:cor_sen}) to artificially control the amount of repulsion forces and thereby regulate the number of clusters. The choice of $\alpha$ should be tailored to the specific task at hand. As shown in \cref{fig:rev}, different $\alpha$ values yield different solutions for various tasks.


\section{Conclusion}
The study presents DeepCut, an innovative unsupervised segmentation technique that combines classical graph theory, Graph Neural Networks (GNNs), and self-supervised pre-trained networks. DeepCut's effectiveness is demonstrated by achieving superior performance in object localization, object segmentation, and semantic part segmentation tasks, surpassing existing state-of-the-art methods in accuracy and speed.

The proposed graph structure incorporates node features, allowing more information to be utilized during the clustering process. Consequently, implicit semantic part segmentation is achievable, eliminating the need for post-processing methods. The versatility of this methodology extends to various downstream tasks, such as video segmentation and image matting. Additionally, the study emphasizes the importance of selecting an appropriate combination of unsupervised training method and clustering method, as discussed in Section \cref{sec:deep_features}.

We demonstrate that our method is capable of optimizing various loss functions derived from classical graph theory, including those that are challenging to optimize using conventional tools (e.g. correlation clustering).

\ificcvfinal
\paragraph{Acknowledgements:} 
The authors would like to thank Shir Amir for her useful comments.
This research was supported by the ISRAEL SCIENCE FOUNDATION
(grant No. 3805/21), within the Israel Precision Medicine Partnership program, and by the
European Research Council (ERC) under the European Union’s Horizon 2020 research and innovation programme (grant agreement No. 101000967).
This project also received funding from the Carolito Stiftung.
Amit Aflalo was supported by the Young Weizmann Scholars Diversity and Excellence program.
Dr. Bagon is a Robin Chemers Neustein AI Fellow.
\fi

{\small
\bibliographystyle{ieee_fullname}
\bibliography{egbib}
}

\clearpage
\setcounter{section}{0}

\twocolumn[{
\renewcommand\twocolumn[1][]{#1}
\maketitle
\centering
\LARGE{\emph{Supplementary Material}}
\vspace{18pt}
}]

 \begin{figure}[t]
\includegraphics[width= \linewidth]{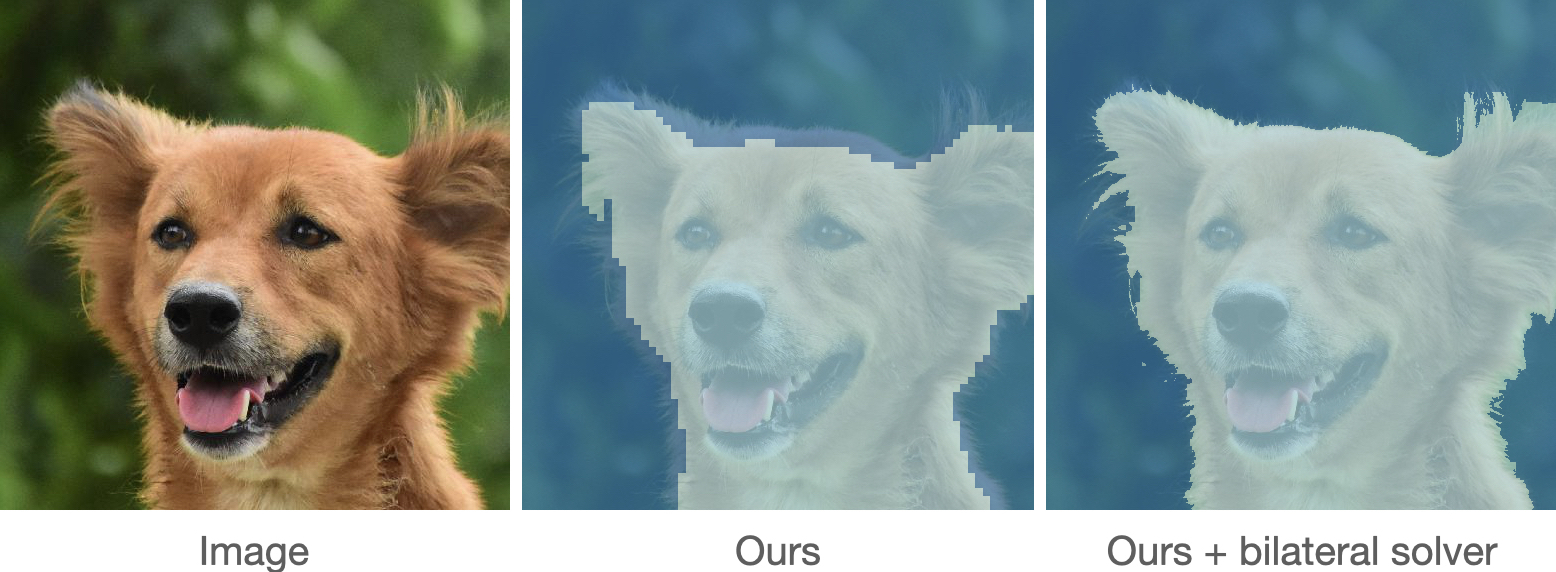}
\caption{
\textbf{Segmentation refinement.} Single object segmentation with our method and bilateral solver\cite{barron2016fast} as a complementary step to refine the obtained segmentation.
}
\label{fig:bs}
 \end{figure}

\begin{figure}[t]
\includegraphics[width= \linewidth]{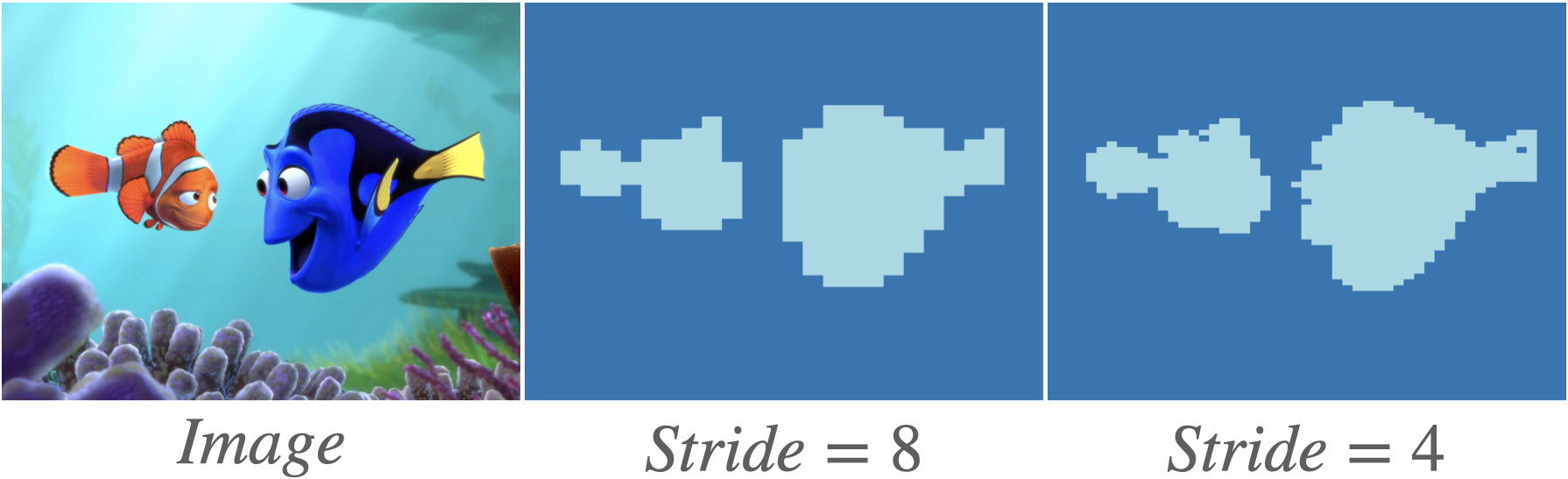}
\caption{
\textbf{Segmentation refinement.} Resolution manipulation using smaller size stride. ViT image input resolution is $280 \times 280$ for both images.
}
\label{fig:nemo}
 \end{figure}
 
\section{Resolution Manipulation}\label{res_man}
In this work, we utilize deep features extracted from ViTs. Those features represent the image patches corresponding to the ViT patch size $p \times p$.
We perform our segmentation patch-wise thus, for image size $(m\times n)$ our segmentation resolution is $(\frac{m}{p} \times \frac{n}{p})$. For higher resolution segmentation, we change the ViT stride value to $\frac{p}{2}$ instead of $p$ as it doubles the number of patches the ViT uses, doubling the resolution of our segmentation to 
$(\frac{2m}{p} \times \frac{2n}{p})$.
We found this method to yield better than changing input image resolutions as the transformer in use in this paper wasn't trained on high resolution images.
Example at \cref{fig:nemo}

In order to improve the segmentation resolution further, a bilateral solver\cite{barron2016fast} can be added as a complementary step to refine the boundaries of the obtained segmentation: see \Cref{fig:bs}. All reported results int the paper is without any post-processing methods.

\section{Implementation details}
\label{sec:intro}
For all experiments, we use DINO \cite{caron2021emerging} trained ViT-S/8 transformer for feature extraction; specifically, we use the \emph{keys} features from the last layer of the DINO trained "student" transformer.  We use pre-trained weights provided by the DINO paper authors (trained on ImageNet\cite{ILSVRC15}).
 \textbf{We do not conduct any training of the transformer on any of the tested datasets.} Input images resized to a resolution of $280 \times 280$, images resized using \emph{Lanczos} interpolation. All expirements where conducted using \emph{Tesla V100} GPU.

\begin{algorithm}
\caption{DeepCut}\label{alg:cap}
\begin{algorithmic}[1]
\State $x \gets Input \ image$
\State $x \gets ViT(x)$ \Comment{Deep features from ViT}
\State $G \gets Build \ graph(x)$
\For{Each training epoch} 
    \State $s \gets GCN(G)$ \Comment{Single layer of GCN}
    \State $s \gets ELU(s)$
    \State $s \gets MLP(s)$ \Comment{Two layer MLP}
    \State $s \gets softmax(s)$
    \State $Loss \gets \mathcal{L}{NCut}$ or $\mathcal{L}{CC}$
\EndFor
\State $Output \ segmentation \gets argmax(s)$
\end{algorithmic}
\end{algorithm}

\paragraph{ViT} Evaluation mode, frozen weights.

\paragraph{Build graph} Create a graph from deep features as depicted at \cref{sec:deep_to}.

\paragraph{GCN} Graph Convolutional Network\cite{kipf2016semi}. \emph{Input\ size} = Deep features size, \emph{hidden size} = 64.

\paragraph{ELU} The Exponential Linear Unit activation function.
\paragraph{MLP} Consists from 2 linear layers, \textbf{layer 1:} from GCN \emph{hidden size} to $\frac{hidden\ size}{2}$. \textbf{layer 2:} from $\frac{hidden\ size}{2}$ to $k$ the number of desired clusters. For k-less usage with correlation clustering, the output will be set to a maximum of desired clusters. Between the layers, there is an \emph{elu} activation function and 0.25 dropout.

 \paragraph{Loss} We suggest two loss function derived from classical graph theory; NCut and CC.

\paragraph{Output segmentation} At step 8, in order to obtain the final segmentation, the vector $s$ is extracted. Each entry in this vector corresponds to a patch of the image and contains a probability vector that describes the likelihood of the patch belonging to a specific cluster. We chose the most likely cluster assignment for each patch and than unflatten the result to get an segmentation map. 

\subsection{Two-stage segmentation}
\begin{figure*}[ht]
\centering
\includegraphics[width=17cm]{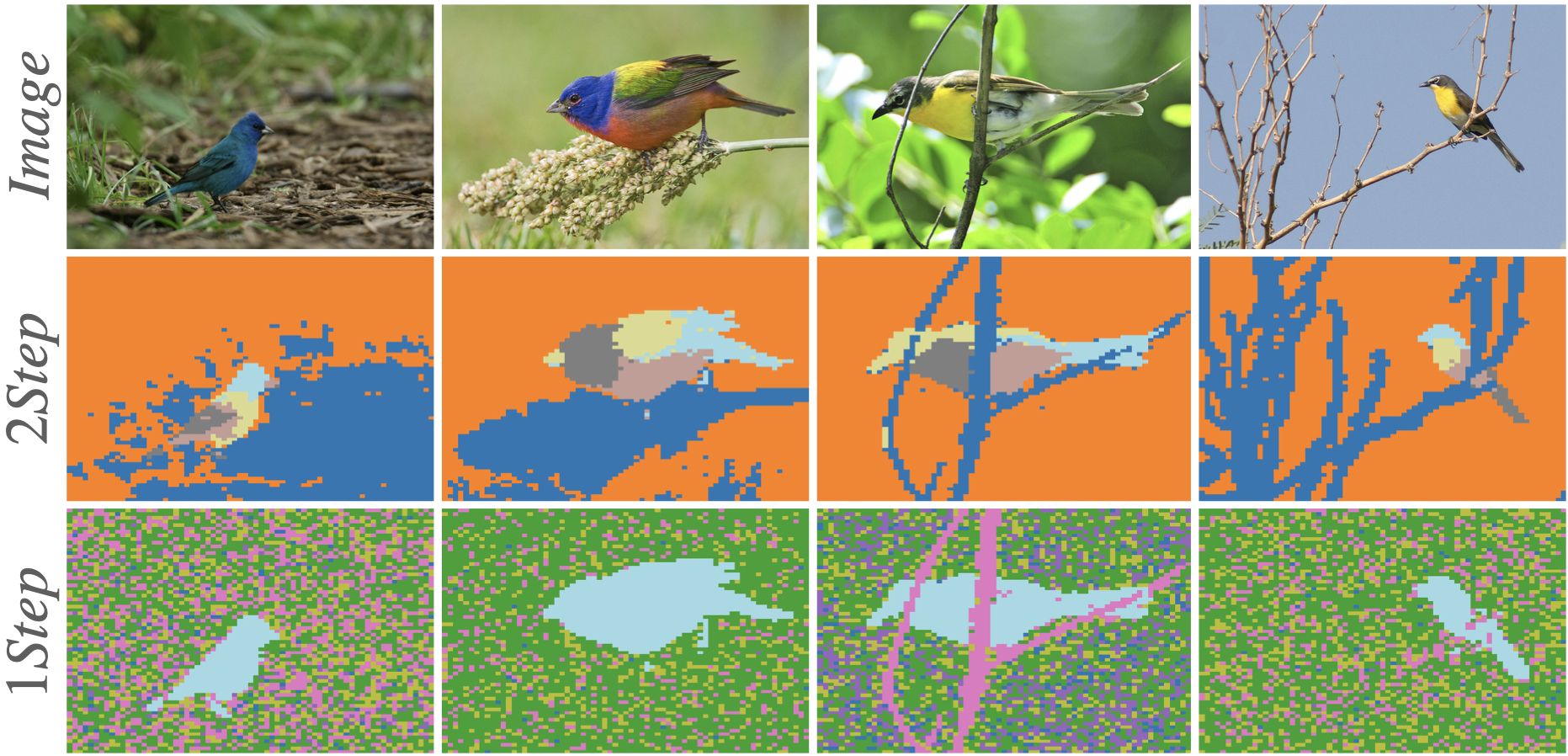}
\caption{
\textbf{Method example:} Two-stage segmentation using DeepCut(NCut).
}
\label{fig:2stp}
\end{figure*}

The clustering functionals in this paper exhibit are biased towards larger clusters (e.g background-foreground), resulting in a tendency to underperform on finer details by merging them together, or in some cases, failing and introducing a significant amount of noise to the segmentation process. To address this issue, a solution is proposed by utilizing a two-stage segmentation approach, where the background and foreground segments are separately applied to avoid the aforementioned biases and limitations. Example can be seen at \cref{fig:2stp}.

\subsection{Training}
To optimize object localization and object segmentation task, we perform individual optimization for each image for a duration of 10 epochs, with model weights being reset between images. For part semantic segmentation, we carry out separate optimization for each image over a span of 100 epochs, without resetting the model weights between them.

\subsection{Performance}
All of the results presented in the paper demonstrate DeepCut without the utilization of any post-processing (e.g. bilateral solver). All experiments were conducted using the same hardware: Tesla V100 GPU and an Intel Xeon 32 core CPU.

\noindent
\begin{tabular}{|l|c|c|c|}
\hline
            & DUTS & ECSSD & Throughput \\
  Model  & [mIoU] & [mIoU] & [img/sec] \\ \hline
TokenCut + Bilateral Solver & 62.4 & 77.2 & 0.5 \\ \hline
TokenCut w/o Bilateral Sol. & 57.6 & 71.2 & 1 \\ \hline
Ours          & 59.5 & 74.6  & 5 \\ \hline
\end{tabular}

\begin{figure*}[ht]
\centering
\includegraphics[width=17cm]{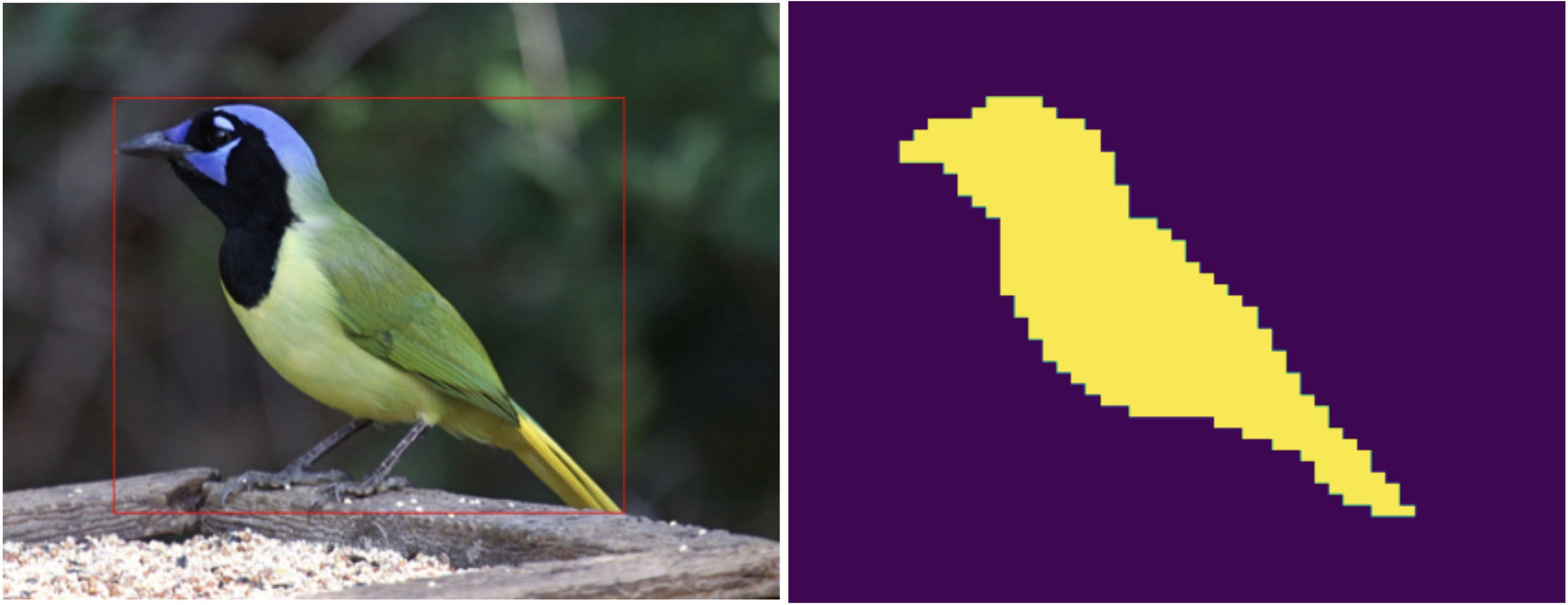}
\caption{
\textbf{Method example:} Object localization using DeepCut(NCut).
}
\label{fig:obloc}
\end{figure*}

\begin{figure*}[ht]
\centering
\includegraphics[width=17cm]{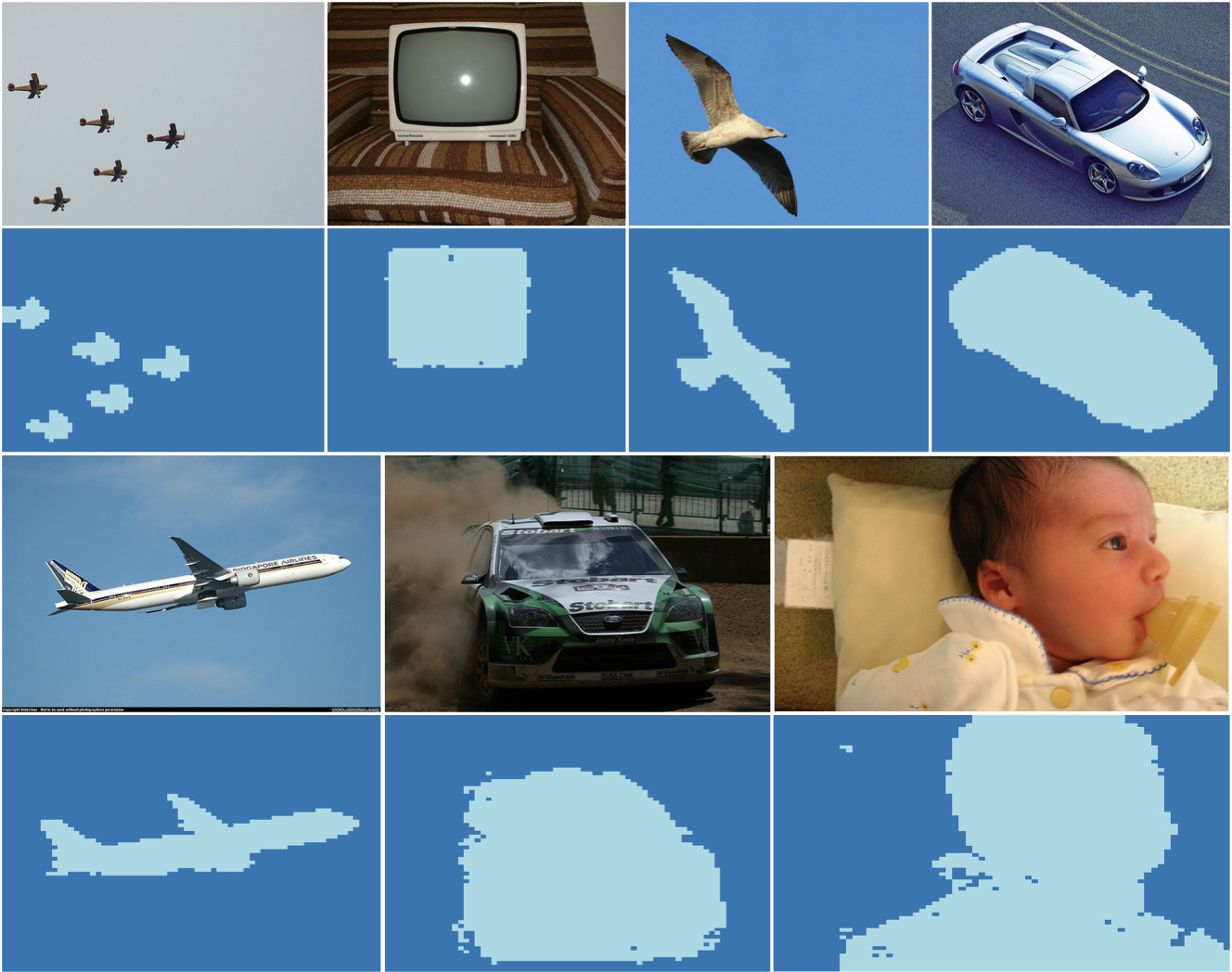}
\caption{
\textbf{Method example:} Random foreground-background segmentation samples using DeepCut(NCut) on VOC07.
}
\label{fig:seg}
\end{figure*}

\begin{figure*}[ht]
\centering
\includegraphics[width=17cm]{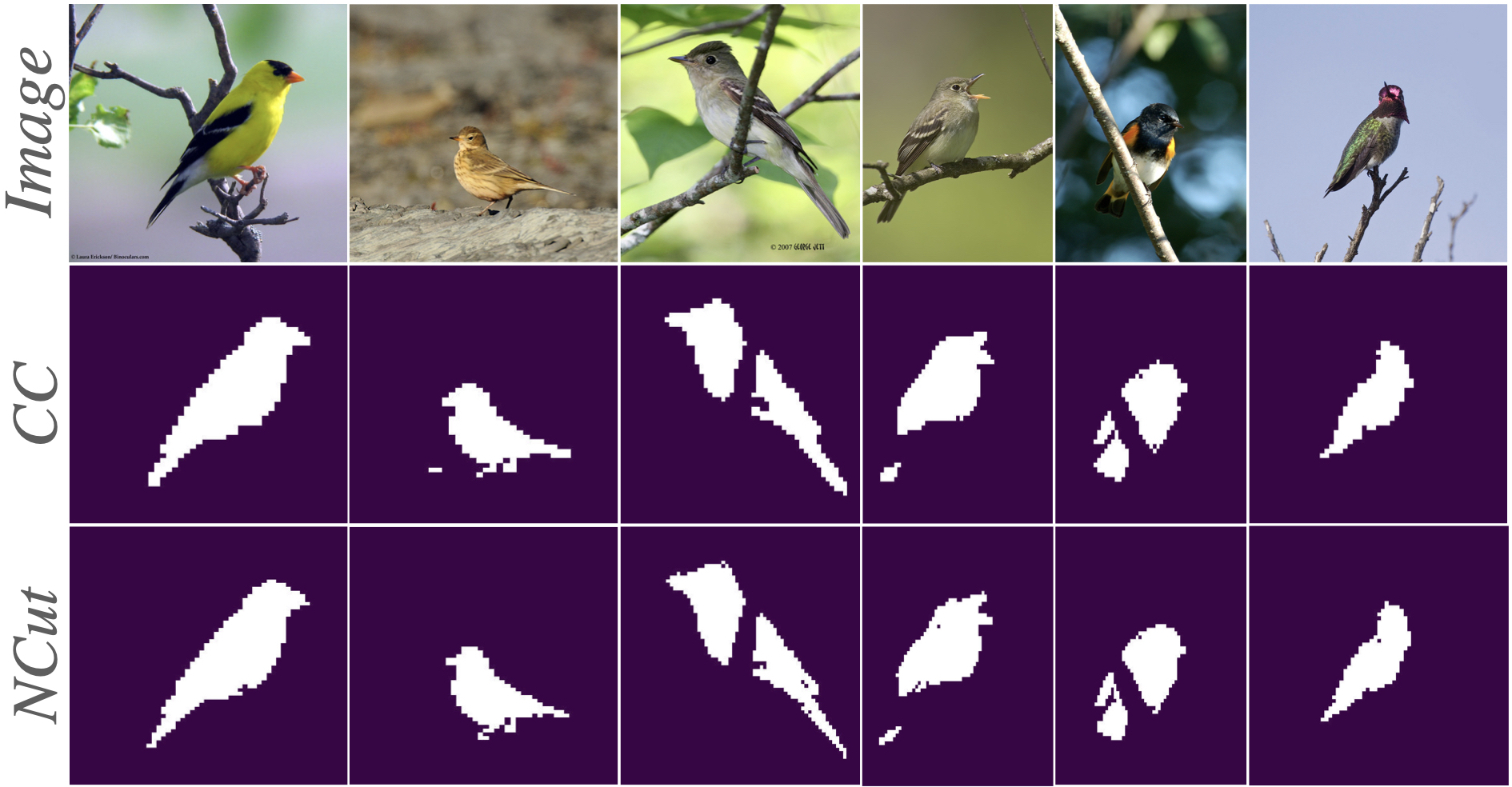}
\caption{
\textbf{Method example:} Random foreground-background segmentation samples using DeepCut(NCut/CC) on CUB-200. DeepCut segments the birds accurately without including other objects such as branches and leaves (which is a common failure point of previous methods).
}
\label{fig:ccncut}
\end{figure*}


\end{document}